\documentclass{article}

\usepackage[nonatbib]{neurips_2021}

\usepackage[utf8]{inputenc} 
\usepackage[T1]{fontenc}    
\usepackage{hyperref}       
\usepackage{url}            
\usepackage{booktabs}       
\usepackage{amsfonts}       
\usepackage{nicefrac}       
\usepackage{microtype}      
\usepackage{xcolor}         
\usepackage{graphicx}
\usepackage{listings}
\usepackage{float}
\usepackage{enumitem}

\title{Comparing Learning Paradigms for Egocentric Video Summarization}
\author{Daniel Wen}
\date{\today}

\begin{document}

\maketitle

\begin{abstract}
  In this study, we investigate various computer vision paradigms—supervised learning, unsupervised learning, and prompt fine-tuning—by assessing their ability to understand and interpret egocentric video data. Specifically, we examine Shotluck Holmes (state-of-the-art supervised learning), TAC-SUM (state-of-the-art unsupervised learning), and GPT-4o (a prompt fine-tuned pre-trained model), evaluating their effectiveness in video summarization. Our results demonstrate that current state-of-the-art models perform less effectively on first-person videos compared to third-person videos, highlighting the need for further advancements in the egocentric video domain. Notably, a prompt fine-tuned general-purpose GPT-4o model outperforms these specialized models, emphasizing the limitations of existing approaches in adapting to the unique challenges of first-person perspectives. Although our evaluation is conducted on a small subset of egocentric videos from the Ego-Exo4D dataset\cite{egoexo4D} due to resource constraints, the primary objective of this research is to provide a comprehensive proof-of-concept analysis aimed at advancing the application of computer vision techniques to first-person videos. By exploring novel methodologies and evaluating their potential, we aim to contribute to the ongoing development of models capable of effectively processing and interpreting egocentric perspectives.

\end{abstract}


\section{Introduction}

First-person videos, particularly those captured by wearable devices such as GoPros, augmented reality glasses, and mobile phones, are becoming increasingly prevalent in everyday life. The growth of platforms like YouTube and social media has further fueled the rise of first-person content, where individuals share personal experiences. While advances in machine learning and artificial intelligence have greatly enhanced computer vision capabilities in third-person video analysis—particularly in tasks such as object segmentation, action recognition, and event detection—understanding egocentric video remains relatively under-explored. Current state-of-the-art models for third-person video analysis often struggle with first-person videos, which present unique challenges such as severe camera shake, limited field of view, and occlusions that obscure critical contextual information. Additionally, while third-person videos are filmed from an external observer's perspective, egocentric videos capture a subjective viewpoint marked by dynamic camera movements, personal interactions, and an inherent focus on the individual’s perspective. These differences make it difficult to generalize actions and interactions across diverse scenes and objects in first-person videos.

\begin{figure}[H]
    \centering
    \makebox[\textwidth][c]{%
    \includegraphics[width=1\textwidth]{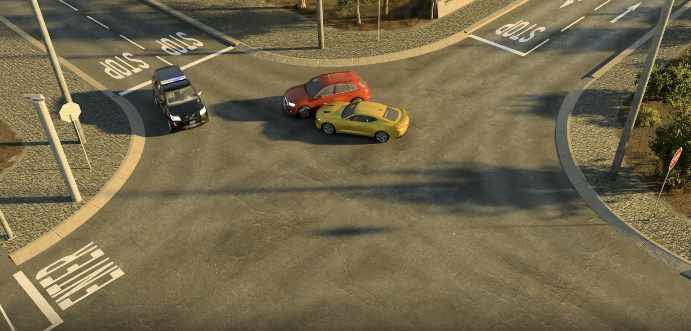}
    }
    \caption{Nvidia Video Summarization Example}
    \label{fig:nvidia_example}
\end{figure}

This frame is taken from an example video produced by Nvidia's latest video summarization agent \cite{nvidia}. The model demonstrates strong performance in video summarization and narration on third-person videos. However, it is crucial to acknowledge that these results were obtained in a controlled, simulated environment with limited variables and no camera movement, conditions that contribute to more effective summarization. In contrast, egocentric videos are inherently more unpredictable, marked by dynamic camera movements and frequent shifts in perspective, which introduce additional challenges for summarization.

This paper compares unsupervised learning, supervised learning, and prompt fine-tuning paradigms, evaluating their performance on egocentric videos. In doing so, we also investigate the unique challenges posed by first-person videos in contrast to third-person videos. Prompt fine-tuning involves iteratively refining prompts to optimize pre-trained language models for specific tasks. In contrast, unsupervised learning trains models on unlabeled data, aiming to uncover inherent patterns or structures within the data. Supervised learning, by contrast, relies on labeled input-output pairs, enabling models to learn a mapping from inputs to outputs and minimize prediction errors on unseen data.

Acknowledging our resource limitations, the objective is not to develop a state-of-the-art solution, but rather to prove contemporary models are insufficient for egocentric video summarization, thereby contributing to the advancement of computer vision models in the egocentric video domain.

\section{Related Work}

Note: Further implementations of each model are detailed in the Methodology section.

\subsection{Shotluck Holmes}

Shotluck Holmes\cite{shotluck} employs advanced pretraining and data collection strategies to enhance the capabilities of small-scale lightweight vision models (LLVMs), enabling them to transition from interpreting single frames to comprehending sequences of frames. They achieve state-of-the-art results on Shot2Story video captioning and summary task, utilizing models that are both significantly smaller and more computationally efficient. Specifically, they integrate the Shot2Story20K paper's findings with TinyLLaVA, one of the leading small-scale multi-modal model families. They demonstrate that substituting the final two stages of Shot2Story20K's three-stage vision-language model pipeline with TinyLLaVA significantly reduces memory consumption, computational requirements, and latency. For multi-shot video summarization, they sample the video frames, feed them into an video encoder to produce visual tokens, feed the tokens into an MLP, concatenate them to an LLM with a text prompt, and then they feed the tensor through a small-scale LLM to generate the caption for the video shot. We use Shotluck Holmes in our evaluation as the state-of-the-art supervised learning model for video summarization.

\subsection{TAC-SUM}

TAC-SUM\cite{tacsum} is an unsupervised video summarizing model that incorporates temporal context to improve cluster-based models. They partition the input video into segments and use their clustering information as context for temporal awareness in the clustering process. They compute the final video summary using the temporal-aware clusters and achieve state-of-the-art on SumMe\cite{summe}. Similarly, we  generate embeddings from the frames of each video of the Ego4D dataset and follow their architecture for video summarization. We also take inspiration from TAC-SUM's temporal context, implementing a similar technique in our prompt fine-tuned method. They even have performance results of TAC-SUM w/o TC (temporal context) which had worse results. We use TAC-SUM in our evaluation as the state-of-the-art unsupervised learning model for video summarization.

\subsection{GPT-4/GPT-4o}

GPT-4\cite{gpt4} is a large-scale, multimodal transformer-based model, pre-trained and fine-tuned to perform a wide range of natural language processing tasks, such as summarization, reasoning, and generation. While GPT-4 is pre-trained to predict the next token in a text sequence, it's successor, GPT-4o extends these capabilities to video understanding. GPT-4o is designed to not only comprehend textual information but also analyze visual content, making it highly proficient at interpreting video data. We take advantage of GPT-4o's multimodal versatility by prompt fine-tuning it for video summarization and comparing it's results to state-of-the-art video summarization models that are fine-tuned specifically on egocentric videos.

\subsection{Ego Exo4D}

Ego-Exo4D\cite{egoexo4D} is one of the few diverse, well-annotated first-person video datasets, so it is heavily utilized within this paper for evaluation. Ego-Exo4D provides a multimodal, multiview video dataset of both egocentric and exocentric videos. Specifically, they capture human activities, such as sports, music, dance, and bike repairs, in videos of lengths ranging from 1 to 42 minutes. The video is accompanied by multichannel audio, camera poses, 3D point clouds, and multiple paired language descriptions. They also provide their own suite of benchmark tests and annotations. It is worth emphasizing that, since we are using this dataset for evaluation purposes, it cannot be used for training; thus, we utilize a comparable yet distinct dataset, Ego4D\cite{ego4D} instead. 

\subsection{Shot2Story20K}

Shot2Story\cite{shot2story} is a multi-shot video understanding benchmark with detailed caption and video summary annotations, sourced from 20,000 videos of the public video benchmark, HDvila100M\cite{hdvilla}. They split each video into shots using a shot detection method, TransNetV2\cite{transnet}, and then they annotate the video summaries based on the shot captions. Shot2Story provides captions for both human narrations and visual signals in order to enhance video comprehension. Furthermore, they design several tasks including video retrieval with shot descriptions, multi-shot video summarization, and single-shot video and narration captioning. Shotluck Holmes' original architecture is fine-tuned on this dataset and achieves state-of-the-art results.

\subsection{UniVTG}

UniVTG\cite{univtg} by Show Lab, Meta AI, and Johns Hopkins University, is a video temporal grounding (VTG) unification method. Video temporal grounding focuses on aligning target segments of videos with custom language queries; however, most existing approaches rely on task-specific models trained with labels tailored to particular tasks, such as moment retrieval or highlight detection. Moment retrieval tasks commonly have time intervals and highlight detection tasks have worthiness curves, but UniVTG unifies the VTG labels and tasks in three procedures: First, they just review a broad spectrum of VTG labels and tasks, proposing a unified framework for their formulation. They then build on this by designing scalable data annotation schemes to generate pseudo-supervision. Second, they introduce a grounding model that can tackle each task and fully leverage the available labels. Third, by utilizing the unified framework, they enable temporal grounding pretraining with large-scale, diverse labels, enhancing their model's grounding capabilities, including zero-shot grounding. 

\subsection{VideoLLM-online}

VideoLLM-online\cite{videollm} is another approach to video understanding that focuses on achieving streaming voice-driven multimodal assistance. More specifically, they aim to generate real-time answers of video streams which continuously refreshes the visual context. This task comes with challenges: First, the user query may come with temporally aligned requirements, such as if the user asks the model to notify them when an action occurs. Second, the model must retain a large amount of vision and language context for summarization. Third, their model must perform as well as other video understanding models while also providing the real-time query answering. They propose Learning-In-Video-strEam (LIVE), a framework that incorporates learning, data, and inference methods to develop an online video assistant. LIVE offers a streaming dialogue generation scheme that enables free-form chatting by converting offline annotations into online dialogues. Doing so allows them to provide data from user queries and assistant responses, which is often scarce in popular video datasets. LIVE also uses continuous key-value caching to aid in the streaming, and parallelizes the visual encoding and language decoding to prevent bottlenecks. This is necessary for real-time application, because visual encoding is quick while language decoding is slow. VideoLLM-online uses CLIP for their vision encoder and Llama-2/Llama-2 for their language model, and their experiments on real-time Ego4D narration shows that their architecture shows advantages in speed and memory cost. However, their model can only sustain such results on video streaming no longer than 5 minutes, and they leave these results as the stepping stone for future real-world usage. 
Our GPT implementation in this paper takes inspiration from VideoLLM-online's focus on video narration to train and evaluate their model's video understanding. Furthermore, we also analyze the correlation between high quality video narrations and the model's performance on temporal action localization tasks.

\section{Methodology}

We will now describe the implementations of GPT-4o for prompt fine-tuning, TAC-SUM for state-of-the-art unsupervised learning, and Shotluck Holmes for state-of-the-art supervised learning. Subsequently, their performance in video summarization on egocentric videos from the Ego-Exo4D dataset will be compared in the Results section.

\begin{lstlisting}[caption={Chain-of-Thought Prompting}, label={lst:CoT}]
"role": "system",
"content": f"""
You are given the frames of an egocentric video. Egocentric videos tend to have redundant or less relevant information. Summarizing these videos means identifying the most
visually or contextually important moments. Return an output for each step. If there is a question in a step, output the response.

Step 1. Segment the video into distinct activity intervals by analyzing motion changes (e.g., sudden bursts of motion) and detecting scene changes based on visual 
context shifts, background changes, or other environmental cues.

Step 2. Key activities can typically be defined by interactions with objects, people, or changes in the environment.
Within each identified activity interval from Step 1, determine what the key activity is by analyzing motion patterns, object presence, and environmental changes.
What are the most significant activities in each segment?

Step 3. Analyze these key activities and remove segments that are repetitive or less relevant to the video.
The goal is to ensure the remaining representative intervals align with important contextual changes, transitions, or motion patterns.
Which activities remain? Which were removed?

Step 4. Combine these selected intervals into a coherent and chronologically ordered summary timeline that maintains the flow and context of the original video 
while emphasizing the most critical moments.

"""
    
\end{lstlisting}

\subsection{Prompt Fine-tuning}

For prompt fine-tuning, we provide two solutions that utilize GPT-4o's state-of-the-art general-purpose reasoning and natural language processing. First, we examine a solution in which we process the entire video's frame simultaneously to the GPT model, utilizing chain-of-thought prompting\cite{CoT} to improve GPT-4o's performance through multi-step reasoning(\ref{lst:CoT}). We then contrast this solution with a short, concise prompt(\ref{lst:HaC}) and call the API for each individual frame of the video, rather than providing all the frames at once. Additionally, this approach enables us to implement history-aware contextualization and batching. History-aware contextualization involves leveraging prior narrations to provide context for the current frame, and batching refers to processing a fixed number of frames concurrently. These techniques enhance the model's temporal awareness and comprehensive understanding of the activities.

\begin{lstlisting}[caption={History-aware Contextualization}, label={lst:HaC}]
    "role": "user",
        "content": [
            {
                "type": "text",
                "text": "Please generate a concise narration for the following frame based on its timestamp and previous narrations as context.",
            },
            {
                "type": "image_url",
                "image_url": {
                    "url": f"data:image/jpeg;base64,{frame}"
                },
            },
            {
                "type": "text",
                "text": f"Previous narrations:\n{previous_narrations}" if previous_narrations else "No previous narrations."
            },
            {
                "type": "text",
                "text": f"The timestamp for this frame (in seconds) is: {timestamp}"
            },
        ]
\end{lstlisting}

\subsection{TAC-SUM}

\begin{figure}[ht]
    \centering
    \makebox[\textwidth][c]{%
    \includegraphics[width=1\textwidth]{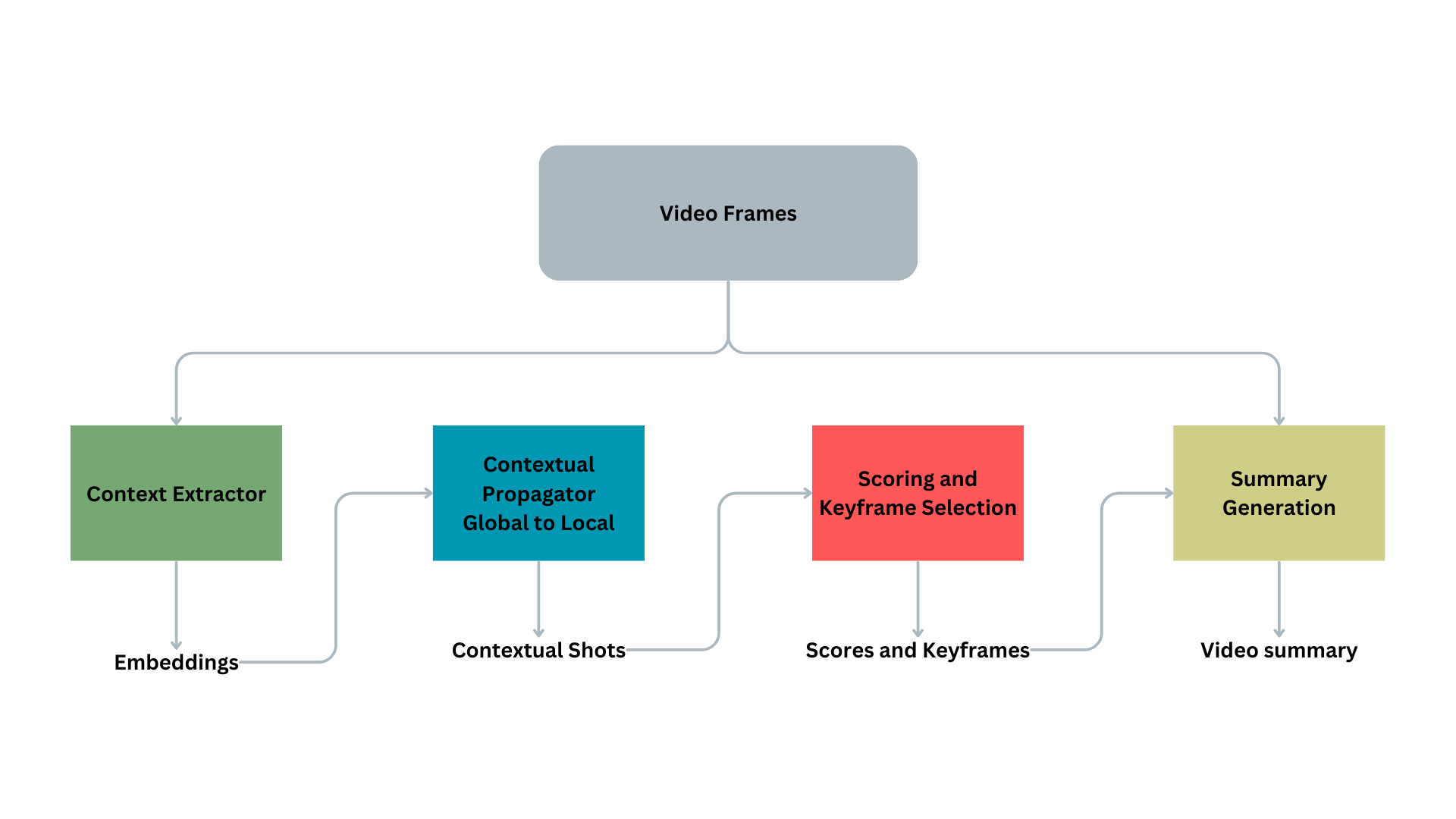}
    }
    \caption{TAC-SUM model architecture}
    \label{fig:tacsum}
\end{figure}

We follow the TAC-SUM paper's implementation. Reference Figure 2 for the following explanation.

First, we generate \textbf{contextual} embeddings with a sampling and embedding step. To reduce computational complexity when processing video data, a \textbf{sampling technique}  is employed to create a smaller, representative sequence of frames from the original video frames, \(I\). The resulting sequence \(\hat{I}\) is sampled with a target frame rate of \(R = 4\) frames per second to ensure consistency and normalization across different video inputs, regardless of their original frame rates. The sampling process involves dividing the video frames within each one-second interval into equal-length segments we call snippets. From each snippet, the middle frame is selected as the representative sample. This method ensures that the reduced sequence effectively captures the essential content of the original video while significantly reducing the number of frames, thereby balancing computational efficiency and data fidelity. For each frame in the sampled sequence \(\hat{I_i}\), the CLIP and DINO feature extractors are used to generate its corresponding \textbf{visual embedding} \(e_i\). Similarly to the TAC-SUM paper, we assess the influence of self-supervised spatial features (DINO) versus richer semantic features (CLIP) on the summarization process. This model, represented as a function \( g : \mathbb{R}^{W \times H \times C} \rightarrow \mathbb{R}^D \), takes the frame \(\hat{I-i}\) with dimensions \(W\) (width), \(H\) (height), and \(C\) (channels, typically color channels) and transforms it into a feature vector of size \(D\). The embeddings from all sampled frames are then concatenated to form the overall contextual embedding of the video, represented as \(E = \{e_1,e_2,...,e_{\hat{T}}\}\). This process captures the contextual features of the video in a compact and structured format.

Second, we break down contextual embedding \(E\) into more detailed, localized representations through \textbf{contextual clustering}  and \textbf{semantic partitioning}. The contextual clustering process is designed to identify relationships between visual elements in the video by grouping frames into meaningful clusters. To achieve this, the high-dimensional contextual embedding \(E\) is first transformed into a lower-dimensional representation, \(\hat{E}\), using dimensionality reduction techniques like PCA and t-SNE from \textbf{scikit-learn}. This reduced embedding facilitates the clustering process by simplifying the data while retaining essential patterns. A two-step clustering approach is then employed. Initially, the BIRCH algorithm\cite{birch} is used to perform coarse clustering, assigning the sampled frames to preliminary groups, represented as \(\hat{c} = \{\hat{c_1}, \hat{c_2},...,\hat{c_{\hat{T}}}\}\). In the next step, hierarchical clustering refines these coarse clusters into finer clusters based on their affinity. The number of final clusters is determined by a sigmoidal function, ensuring that the clustering adapts to the data's complexity while maintaining a pre-set maximum threshold. Each fine cluster is created as a union of one or more coarse clusters, progressively combining similar groups to form semantically coherent clusters. This hierarchical approach allows the embedding \(\hat{E}\) to transition from capturing global patterns to representing finer, localized relationships, ultimately producing clusters that are both meaningful and useful for downstream tasks.

Third, each sampled video frame \(\hat{I_i}\) is assigned a \textbf{cluster label}  \(c_i\) based on its grouping. To ensure the quality of these labels, an outlier elimination step removes frames that deviate significantly from their assigned clusters, while a refinement step merges smaller partitions into larger ones if they fall below a specified threshold \(\epsilon\). A smoothing process is then applied to further refine the labels by assigning each frame a final label \(\hat{c_i}\) determined through majority voting among its neighboring frames. Using these final labels \(C\), the video frames are divided into sections \(P=\{P_1, P_2,...,P_{\hat{N}}\}\), with each section representing a semantically coherent part of the video. These initial partitions are subjected to a refinement algorithm to ensure that each section meets a minimum length requirement \(\epsilon\). The algorithm works iteratively: the shortest partition is identified, and its neighbors are merged with it until the combined length satisfies the threshold. The total number of partitions \(N\) is updated during each iteration, continuing until all partitions are of sufficient length. This semantic partitioning enables the video to be broken into meaningful segments, which can then be independently analyzed. This structure supports detailed analysis and facilitates generating summaries or extracting specific characteristics from the video in later stages.

Fourth, we use the resulting partitions \(P\) to generate \textbf{keyframes} \(k\), a subset of indices of sampled frames \(k \subset t\), and \(k = \cup_{i=1}^{N}\) , where \(k^{(i)}\) are the partition-wise keyframes. The importance of each sampled frame \(\hat{I_i}\) is represented as a vector of scores \(v \in \mathbb{R}^{\hat{T}}\). Initially, the importance score for a frame \(\hat{v_i}\) is set based on the length of the section it belongs to. This baseline score is then refined using a keyframe-biasing approach, which adjusts the values depending on the proximity of frames to keyframes. The biasing method allows for flexibility, offering options to either boost the importance of frames closer to keyframes or lower the importance of others further away. To determine the scores for frames situated between key positions, interpolation methods are applied. Two commonly used techniques for this purpose are cosine interpolation, which produces a smooth, curved transition, and linear interpolation, which results in a straight-line progression. These methods create different patterns of importance distribution across the frames, as illustrated in Figure 4. By assigning these importance scores, the process identifies the most significant frames in the video, enabling their prioritization for inclusion in a concise and meaningful video summary. This method ensures that key moments are emphasized while maintaining an overall balance in representation.

Finally, we leverage the keyframes and their importance scores to generate \textbf{time intervals} for video summarization by identifying the temporal boundaries between significant events or segments. We define the intervals as the time spans between consecutive keyframes. For example, if keyframe \(K_i\) occurs as time \(t_i\) and \(K_{i+1}\) occurs at \(t_{i+1}\), then the interval for that segment is \([t_i,t_{i+1}]\). Keyframes are prioritized based on their importance scores and sorted into chronological order (based on their start times) so that they can be used to trim the original video into the summarized one.

\subsection{Shotluck Holmes}

\begin{figure}[H]
    \centering
    \makebox[\textwidth][c]{%
    \includegraphics[width=2\textwidth]{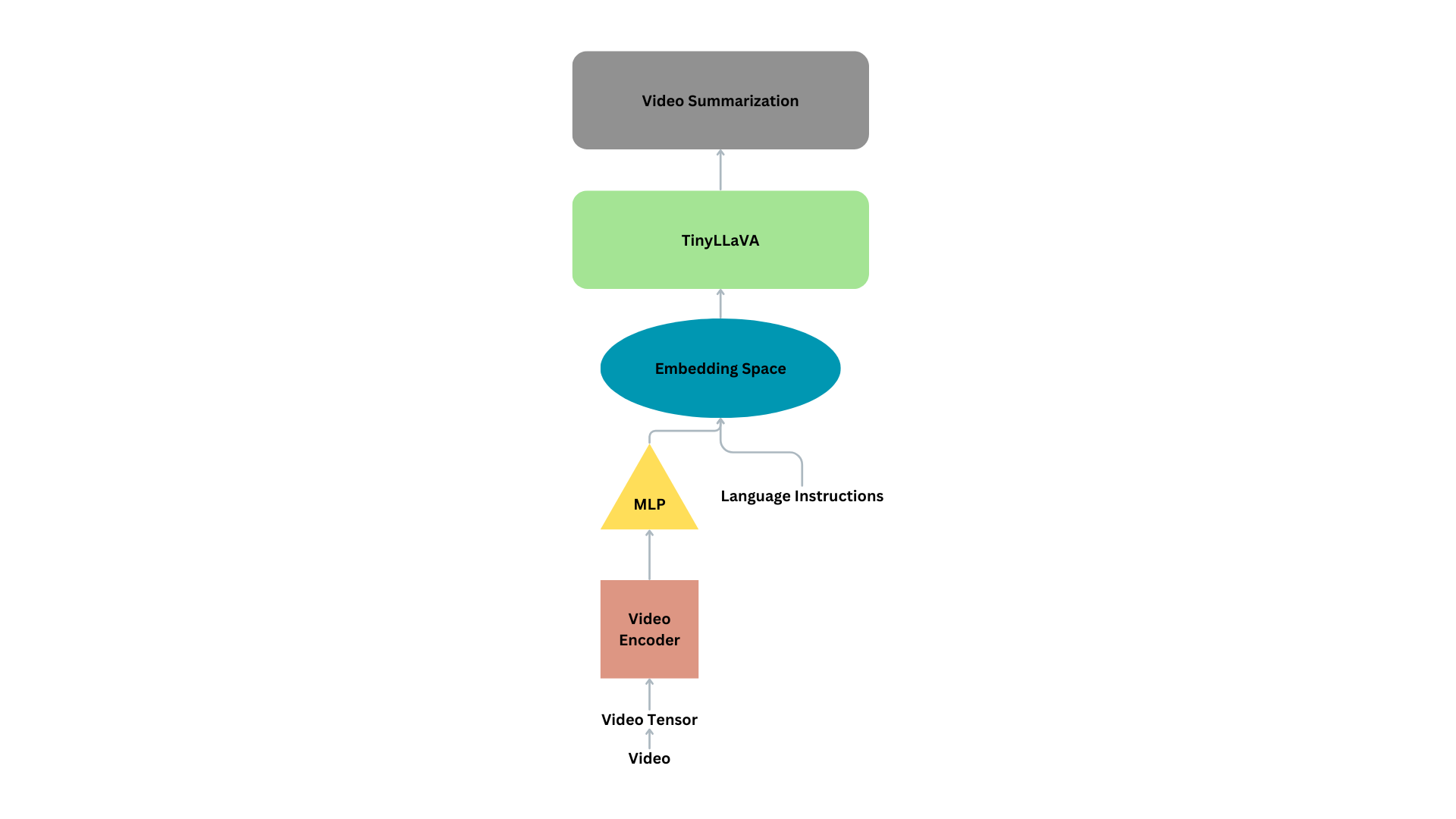}
    }
    \caption{Shotluck Holmes model architecture}
    \label{fig:shotluck}
\end{figure}

To preprocess the Ego4D videos for compatibility with the Shotluck Holmes architecture, each video is converted into a tensor, which is then input into SigLip \cite{siglip}, a vision encoder. To manage the frame limitations of our training hardware, we employ one of two sampling methods: uniform sampling and head-tail sampling. Uniform sampling selects frames at regular intervals across the entire video, while head-tail sampling focuses on the beginning (head) and end (tail) of the video, ensuring an equal number of frames from each half. Following the sampling process, the frames are input into the vision encoder.

The Shot2Story20K model architecture employs multiple efficient small-scale LLVMs to reduce computational complexity in single-shot video captioning and multi-shot video summarization. In contrast, we use a single small-scale LLVM model, TinyLLaVA, which has already been pre-trained with vision encoders. The Shotluck Holmes paper demonstrates that this approach outperforms Shot2Story’s multi-step model by fine-tuning a single, small LLM on the Shot2Story dataset. Similarly, we aim to fine-tune TinyLLaVA on the Ego4D dataset to capture the features of egocentric videos. Video encodings are mapped to the LLM embedding space using a multi-layer perceptron (MLP), and during fine-tuning, we freeze the initial 12 layers, updating the remaining model parameters. A visual representation of the model architecture is provided in Figure 3. We present a 3.1B parameter LLM based on TinyLLaVA, fine-tuned on Ego4D, which generates video summarization intervals for trimming the original video.

\section{Results}

Note: Each referenced video is labeled with a unique video ID, allowing for easy retrieval if needed. The video summaries of these videos can be found in the provided Google Drive in our Github.

\begin{center}
\begin{tabular}{||c c c||} 
 \hline
 Model & Size (\# of Parameters) & Quality Score  \\ [0.5ex] 
 \hline\hline
 Prompt fine-tuned GPT-4o & undisclosed & 64.95  \\ 
 \hline
 TAC-SUM-CLIP & 151.53M & 56.24  \\
 \hline
 TAC-SUM-DINO & 100.76M & 58.43 \\
 \hline
 Shotluck Holmes & 3.1B & 61.19 \\ [1ex] 
 \hline
\end{tabular}
\end{center}

\textbf{Quality Score} represents the mean of all individual video summary scores for the given model. Higher quality score is equivalent to better performance. See below for the scoring criteria.

\subsection{Evaluation Methodology}

For evaluation, we utilize a subset of 21 first-person videos from the Ego-Exo4D dataset for a total of 84 video summaries, which includes a diverse range of activities such as rock climbing, soccer, cooking, bike repair, playing musical instruments (piano, violin, guitar), dancing, administering COVID-19 tests, and performing CPR. This subset comprehensively represents all activities within the Ego-Exo4D dataset. These videos are annotated with ground-truth video summaries and narrations, providing a comprehensive basis for evaluation. Each model generates a summary- a video trimmed to the key activities that the model deems relevant to include. 

To assess each model's performance, a human evaluator unaffiliated with our research watches follows these steps:

\begin{enumerate}[label=\roman*.]
    \item Review the original Ego-Exo4D video in its entirety.
    \item Evaluate the corresponding video summary generated by the model.
    \item Assess the video summary based on the following criteria:
        \begin{itemize}
            \item \textbf{Accuracy:} Evaluate the extent to which the summary correctly represents the content of the original video (100 points).
            \item \textbf{Clarity:} Assess how well the summary conveys the information in an understandable and coherent manner without erratic time interval skips (100 points).
            \item \textbf{Relevance:} Judge the appropriateness of the content included in the summary in relation to the original video, specifically, the model's ability to comprehend the context and accurately identify the key activities (100 points).
        \end{itemize}
    \item Compute the average score for the video summary by summing the points awarded for each criterion and dividing the total by three.
    \item Repeat the previous steps until all video summaries are scored.
    \item Calculate the overall average score (\textbf{Quality Score}) by summing the individual scores of all summaries and dividing by the total number of videos assessed.
\end{enumerate}

Summaries that accurately capture key activities, objects, and environmental details are awarded higher scores. Although our evaluation relies on a smaller dataset and manual testing, we aim to derive meaningful and significant insights within the constraints of our available resources. 

\subsection{TAC-SUM}

As the current state-of-the-art unsupervised learning model for video summarization, we expect TAC-SUM to outperform all existing models, or at least match the performance of Shotluck Holmes. For feature extraction, we utilize CLIP and DINO to generate visual embeddings, referring to these variants as TAC-SUM-CLIP and TAC-SUM-DINO, respectively. The TAC-SUM paper evaluates the impact of self-supervised spatial features (DINO) versus detailed semantic features (CLIP) on the summarization process, concluding that DINO (dino-b16) outperforms CLIP (clip-base-16) on the SumMe dataset.

\begin{figure}[H]
    \centering
    \begin{minipage}{0.45\textwidth}
        \centering
        \includegraphics[width=\linewidth]{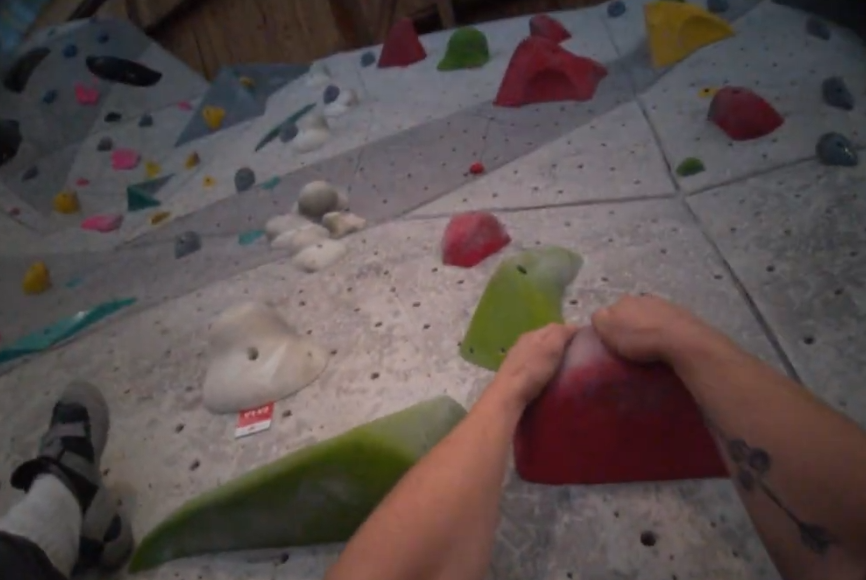}
    \end{minipage}\hfill
    \begin{minipage}{0.45\textwidth}
        \centering
        \includegraphics[width=\linewidth]{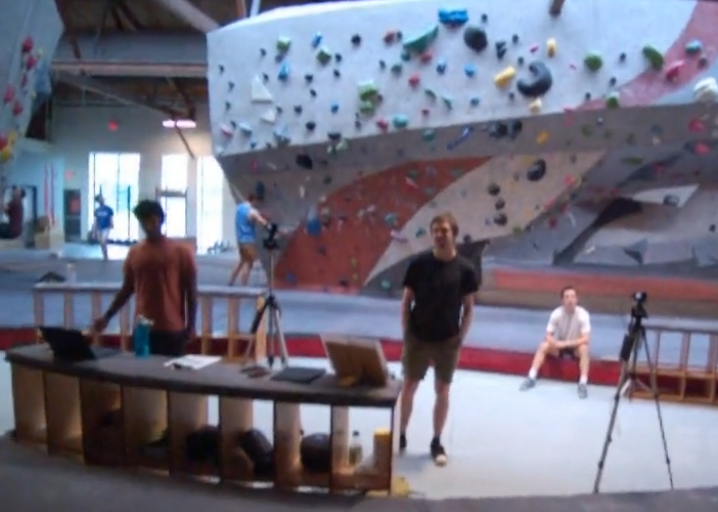}
    \end{minipage}
    \caption{TAC-SUM-DINO vs TAC-SUM-CLIP: Ego-Exo4D minnesota\_rockclimbing\_034\_16 }
    \label{fig:rockclimbing}
\end{figure}

Similarly, we observe that DINO outperforms CLIP on the egocentric test subset from Ego-Exo4D, achieving a quality score of \textbf{58.43} for the DINO model, compared to \textbf{56.24} for the CLIP model. A more detailed analysis of the video summaries reveals that TAC-SUM-DINO is more effective at capturing spatial relationships between different objects. For instance, consider the example of a person rock climbing (\ref{fig:rockclimbing}), where the individual spends most of the video climbing, only pausing briefly to interact with spectators.

The left frame, depicting the person actively rock climbing, is the desired content for inclusion in the video summarization, while the right frame, showing spectators watching the climb, should be excluded. TAC-SUM-DINO focuses on the individual’s hands as they grip the rocks, emphasizing the physical action of climbing. In contrast, TAC-SUM-CLIP highlights the spectators, assuming their relevance to the activity. While it is understandable that TAC-SUM-CLIP identifies the spectators as important—given that people are typically key objects in egocentric activities—TAC-SUM-DINO’s focus on the climbing action demonstrates a more accurate understanding of the primary activity in the video.

This example underscores the distinction between TAC-SUM-DINO and TAC-SUM-CLIP, yet both models exhibit suboptimal performance on egocentric videos, as reflected in their low quality scores. The video summaries reveal that TAC-SUM fails to adeptly interpret spatial and temporal contexts in egocentric videos to identify key actions.

\subsection{Shotluck Holmes}

\begin{figure}[H]
    \centering
    \begin{minipage}{0.45\textwidth}
        \centering
        \includegraphics[width=\linewidth]{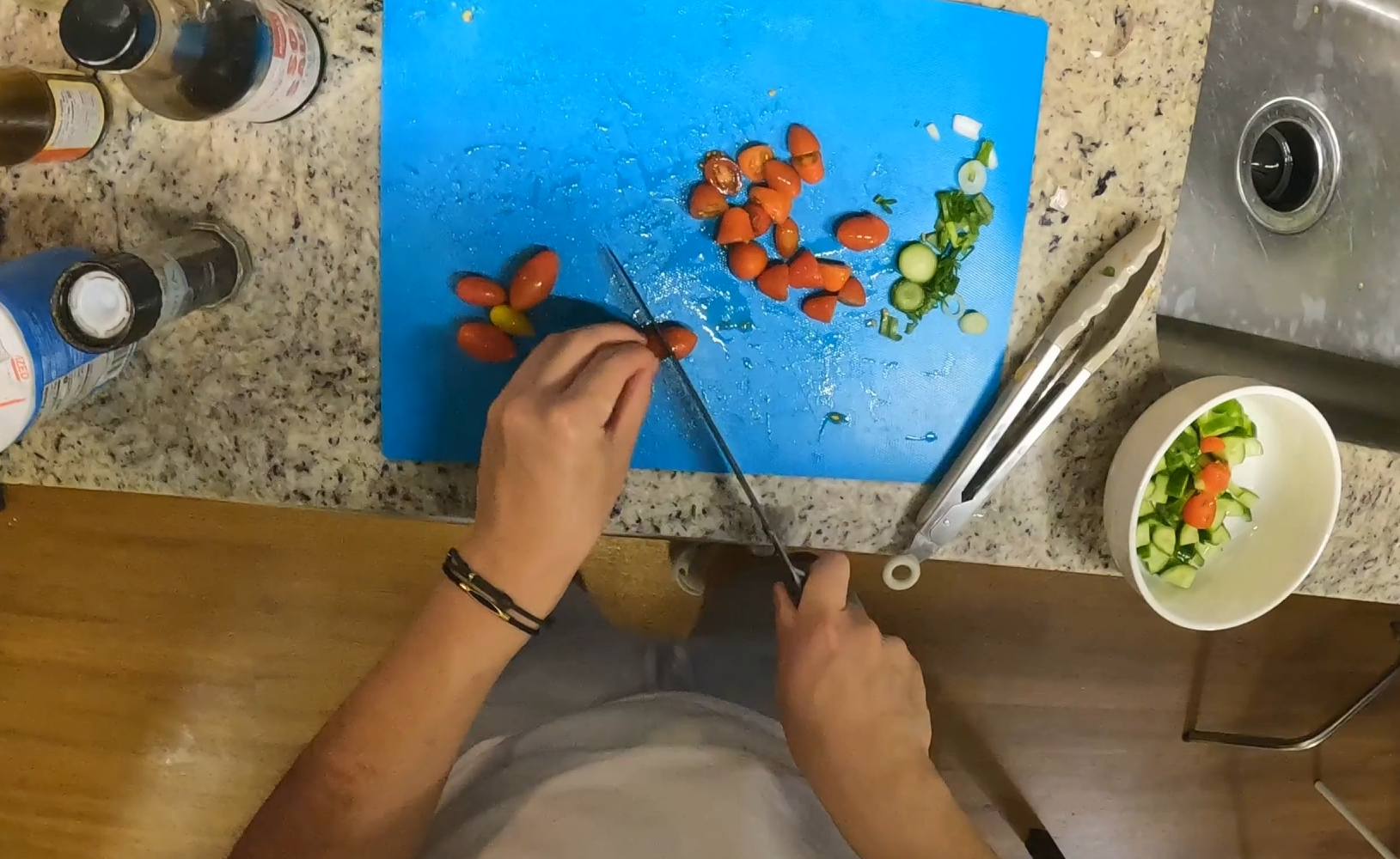}
    \end{minipage}\hfill
    \begin{minipage}{0.45\textwidth}
        \centering
        \includegraphics[width=\linewidth]{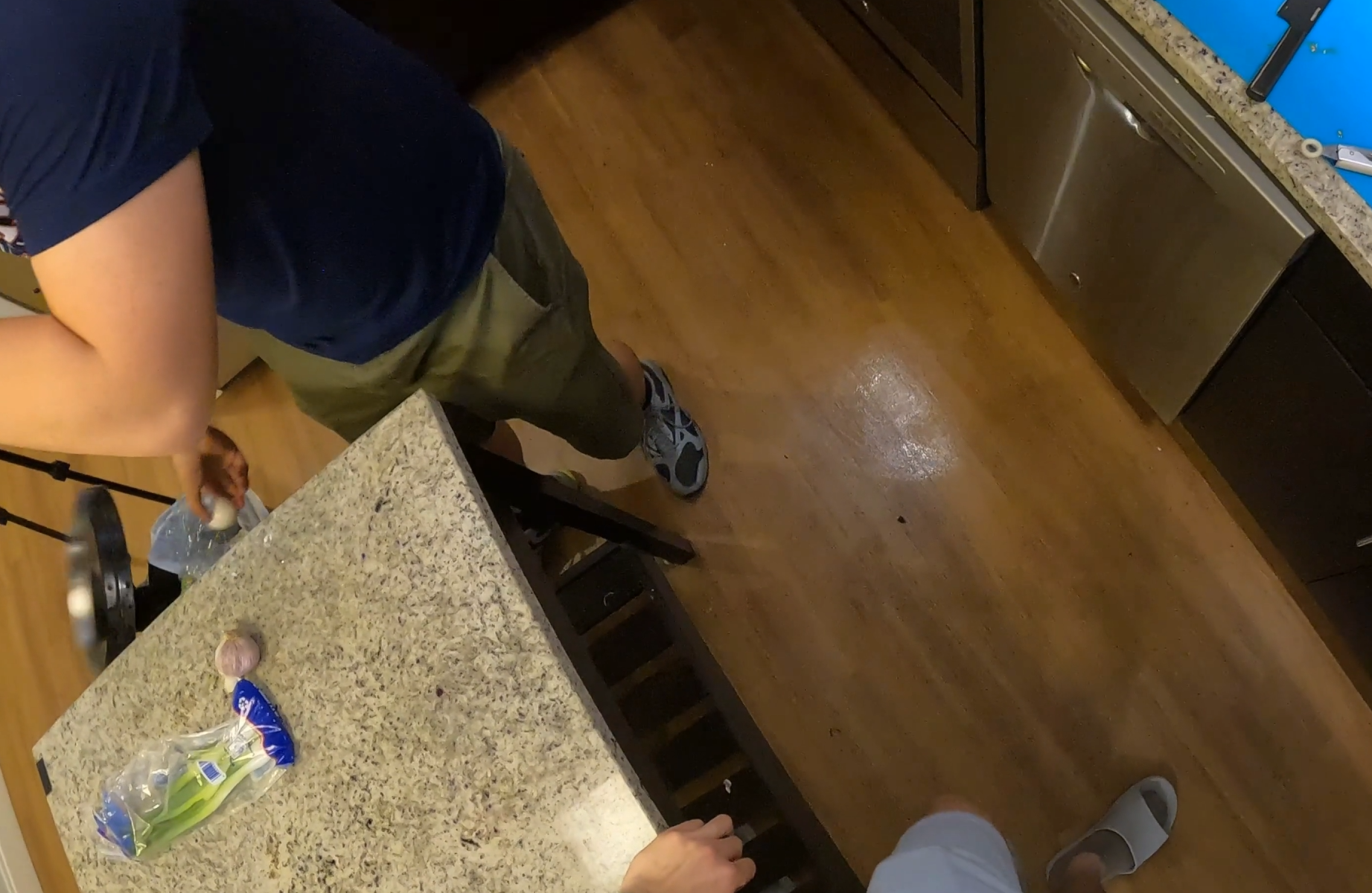}
    \end{minipage}
    \caption{Shotluck Holmes: Ego-Exo4D upenn\_0711\_Cooking\_6\_4 }
    \label{fig:cooking}
\end{figure}

Given TAC-SUM's relatively low quality score, we anticipated that Shotluck Holmes, as the state-of-the-art supervised learning model in video summarization, would outperform TAC-SUM. However, as shown in the results table, Shotluck Holmes achieved a quality score of \textbf{61.19}, which is only marginally higher than that of TAC-SUM. This modest improvement can be attributed to the unique challenges posed by egocentric videos, particularly their dynamic shifts in angles and perspectives, which complicate Shotluck Holmes' ability to contextualize key activities. To illustrate this, we refer to an example of a person preparing a salad by cutting cucumbers and tomatoes, salting them, and adding salad dressing (\ref{fig:cooking}).

The left frame, showing the person cutting tomatoes and cucumbers, represents the content we want the summarization to focus on, while the right frame, depicting the person watching someone else discard eggshells, should be excluded. Ideally, Shotluck Holmes should concentrate solely on actions like those in the left frame, as they are directly relevant to the process of making the salad. However, we observe that it fails to exclude unrelated moments, such as the person interacting with others in the kitchen, clearing the table, retrieving the cucumber from the refrigerator, reading the tomato label, dropping and picking up a tomato, putting items back in the refrigerator, and conversing with others. If Shotluck Holmes correctly identifies salad preparation as the key activity, it should eliminate these extraneous moments, ensuring the summary focuses exclusively on the relevant cooking steps.

\subsection{GPT-4o}

\begin{figure}[H]
    \centering
    \begin{minipage}{0.45\textwidth}
        \centering
        \includegraphics[width=\linewidth]{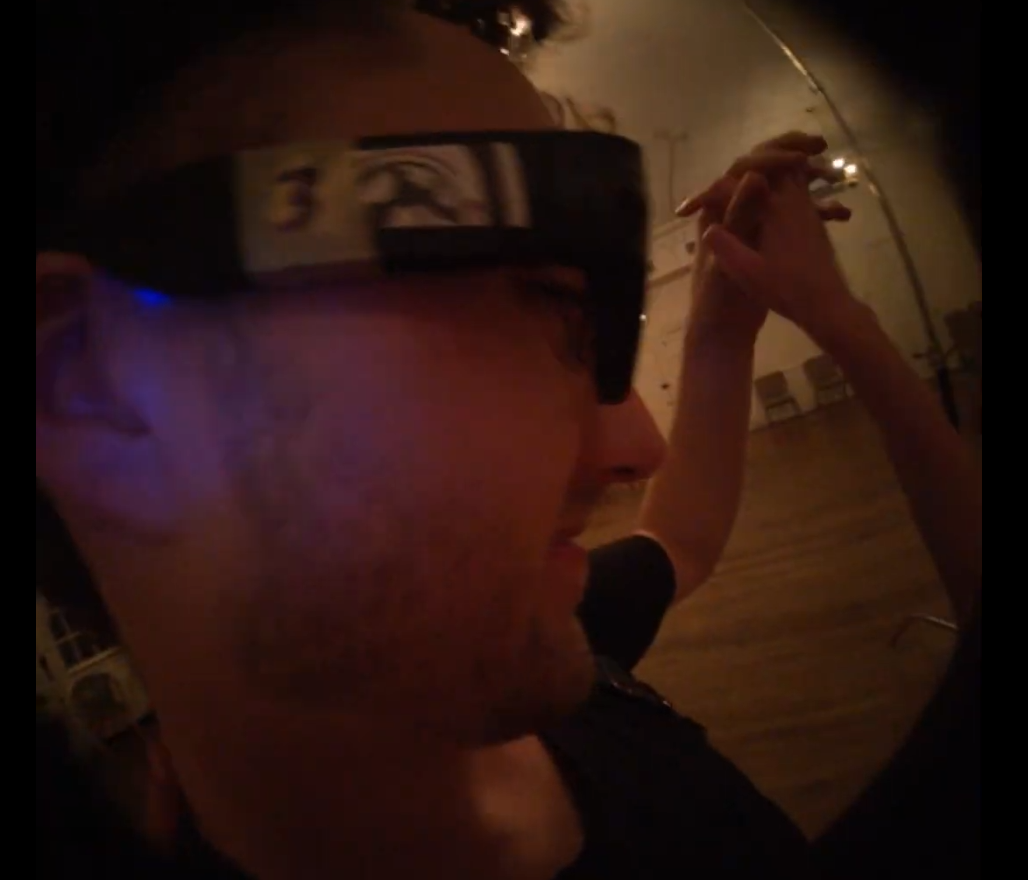}
    \end{minipage}\hfill
    \begin{minipage}{0.45\textwidth}
        \centering
        \includegraphics[width=\linewidth]{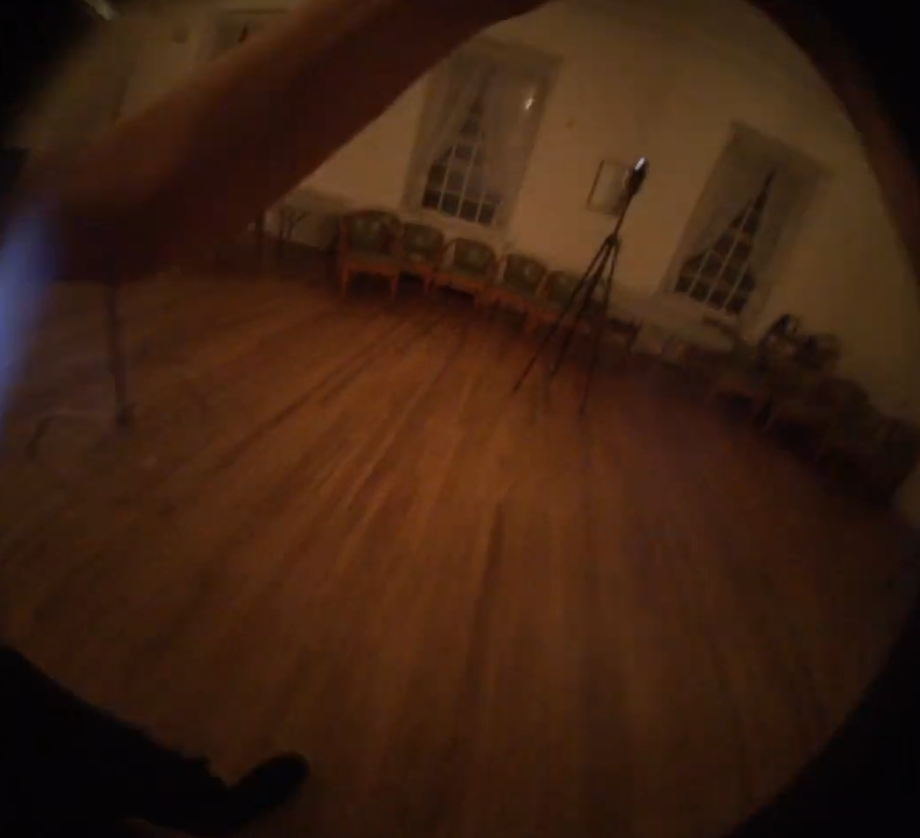}
    \end{minipage}
    \caption{GPT-4o: Ego-Exo4D upenn\_0727\_Partner\_Dance\_2\_2\_5 }
    \label{fig:dancing}
\end{figure}

Having examined the state-of-the-art supervised and unsupervised learning models for video summarization, we now turn to the performance of GPT-4o, fine-tuned using prompt engineering. We initially anticipated its performance would be significantly inferior to the other models; however, with a quality score of \textbf{64.95}, GPT-4o marginally surpasses both TAC-SUM and Shotluck Holmes. In comparison to the rock climbing video discussed in the TAC-SUM section, GPT-4o successfully eliminates interactions with spectators, retaining only the climbing moments. Similarly, in the cooking video referenced in the Shotluck Holmes section, GPT-4o removes most irrelevant actions, such as dropping the tomato. While GPT-4o performs commendably compared to the other state-of-the-art models, its overall performance still demonstrates notable limitations. One significant challenge it faces is handling videos with dynamic camera movements. To illustrate this, we refer to a video of two individuals dancing in a room (\ref{fig:dancing}).

We expect GPT-4o to retain frames like the one on the left, as they are directly relevant to the dance, which is the primary activity in the video. However, the summary produced by GPT-4o exhibits erratic transitions and lacks focus, suggesting that the model struggles to determine which objects or movements to retain or exclude. This issue is particularly evident in egocentric videos with rapid movements, such as the twirl depicted in the right image from the dancing video, where GPT-4o fails to interpret shifts in object perspectives and viewing angles. While the prompt fine-tuned GPT-4o model demonstrates reasonable performance in video summarization overall, its inadequate handling of dynamic camera movements highlights the limitations of this approach for summarizing egocentric videos.

\section{Conclusion}

Our findings successfully evaluates various computer vision paradigms and their performance on egocentric videos. The analysis demonstrates that each learning paradigm exhibits suboptimal performance, with all models encountering significant challenges in addressing the complexities associated with understanding egocentric videos.  Finally, we identify several unique challenges inherent to egocentric videos, such as dynamic camera movements and shifting object perspectives, which explain why existing models—primarily designed for third-person perspectives—struggle with first-person video understanding. Through these findings, we aim to contribute to the advancement of computer vision models by pushing for technological advancement tailored to the egocentric video domain.

\section{Future Work}

As previously discussed, the limited availability of resources necessitated the development of a manual quality ranking system. Future work could include the creation of a larger egocentric video test dataset accompanied by a more systematic and comprehensive evaluation framework. Additionally, evaluating state-of-the-art models across adjacent tasks to video summarization, such as video narration and temporal action localization, would provide deeper insights into video understanding in the context of first-person videos. We also note that Ego-Exo4D focuses on one key activity per video which can be limiting compared to long videos with various unique activities. There is a notable scarcity of robust egocentric video datasets, and the development of such datasets could significantly advance technologies in this domain. While this study does not aim to propose a solution that surpasses all contemporary techniques, addressing the unique challenges identified in egocentric videos could serve this direction for future research.


{
\small

\begin{thebibliography}{9}

\bibitem{univtg} Lin, Kevin Q., et al. "UniVTG: Towards Unified Video-Lanaguage Temporal Grounding." ArXiv (Cornell University), 18 Aug 2023, \url{https://arxiv.org/pdf/2307.16715}

\bibitem{videollm} Chen, Joya, et al. "VideoLLM-online: Online Video Large Language Model for Streaming Video." ArXiv (Cornell University), 17 Jun 2024, \url{https://arxiv.org/pdf/2406.11816}

\bibitem{videomamba} Li, Kunchang, et al. "VideoMamba: State Space Model for Efficient Video Understanding." ArXiv (Cornell University), 12 Mar 2024, \url{https://arxiv.org/pdf/2403.06977}

\bibitem{mamba} Gu, Albert and Dao, Tri. "Mamba: Linear-time Sequence Modeling with Selective State Spaces." ArXiv (Cornell University), 31 May 2024, \url{https://arxiv.org/pdf/2312.00752}

\bibitem{space} Bertasius, Gedas, et al. "Is Space-Time Attention All You Need for Video Understanding?" ArXiv (Cornell University), 9 Jun 2021, \url{https://arxiv.org/pdf/2102.05095}

\bibitem{egoexo4D} Grauman, Dristen, et al. "Ego-Exo4D: Understanding Skilled Human Activity from First- and Third-Person Perspectives." ArXiv (Cornell University), 25 Sep 2024, \url{https://arxiv.org/pdf/2311.18259}

\bibitem{shotluck} Luo, Richard, et al. "Shotluck Holmes: A Family of Efficient Small-Scale Large Language Vision Models for Video Captioning and Summarization." ArXiV (Cornell University), 21 Oct 2024, \url{https://arxiv.org/pdf/2405.20648}

\bibitem{shot2story} Han, Mingfei, et al. "Shot2Story20K: A New Benchmark for Comprehensive Understanding of Multi-shot Videos." ArXiV (Cornell University), 21 Oct 2024, \url{https://arxiv.org/pdf/2312.10300}

\bibitem{hdvilla} Xue, Hongwei, et al. "Advancing High-Resolution Video-Language Representation with Large-Scale Video Transcriptions." ArXiV (Cornell University), 8 Jul 2022, \url{https://arxiv.org/pdf/2111.10337}

\bibitem{transnet} Tom\'{a}\v{s} Sou\v{c}ek and Jakub Loko\v{c}. "Transnet v2: An effective deep network architecture for fast shot transition detection." ArXiV (Cornell University), 11 Aug 2020, \url{https://arxiv.org/pdf/2008.04838}

\bibitem{ego4D} Grauman, Kristen, et al. "Ego4D: Around the World in 3,000 Hours of Egocentric Video." ArXiV (Cornell University), 11 Mar 2022, \url{https://arxiv.org/pdf/2110.07058}

\bibitem{CoD} Adams, Griffin, et al. "From Sparse to Dense: GPT-4 Summarization with Chain of Density Prompting." ArXiV (Cornell University), 8 Sep 2023, \url{https://arxiv.org/pdf/2309.04269}

\bibitem{gpt4} OpenAI. "GPT-4 Technical Report." ArXiV (Cornell University), 4 Mar 2024, \url{https://arxiv.org/pdf/2303.08774}

\bibitem{tacsum} Huynh-Lam, Hai-Dang, et al. "Cluster-based Video Summarization with Temporal Context Awareness." ArXiV (Cornell University), 9 Apr 2024, \url{https://arxiv.org/pdf/2404.04511v1}

\bibitem{summe} Gygli, Michael,et al. "Creating Summaries from User Videos." Computer Vision - ECCV 2014, \url{https://doi.org/10.1007/978-3-319-10584-0_33}

\bibitem{siglip} Zhai, Xiaohua, et al. "Sigmoid Loss for Language Image Pre-Training." ArXiV (Cornell University), 27 March 2023, \url{https://arxiv.org/abs/2303.15343}

\bibitem{birch} Zhang, Tian, et al. "BIRCH: An Efficient Data Clustering Method for Very Large Databases."  SIGMOD Rec. 25(2), 103–114, 1 June 1996, \url{https://doi.org/10.1145/235968.233324} 

\bibitem{CoT} Wei, Jason, et al. "Chain-of-Thought Prompting Elicits Reasoning in Large Language Models." ArXiV (Cornell University), 10 Jan 2023, \url{https://arxiv.org/pdf/2201.11903}

\bibitem{nvidia} Nvidia. Nov 4, 2024, \url{https://build.nvidia.com/nvidia/video-search-and-summarization}

\end{thebibliography}
}
\end{document}